\documentclass[letterpaper]{article} 
\usepackage{arxiv_aaai27}  
\usepackage[hyphens]{url}  
\usepackage{graphicx} 
\usepackage{natbib}  
\usepackage{caption} 
\usepackage{algorithm}
\usepackage{algorithmic}
\usepackage{amssymb}
\usepackage{amsmath}
\usepackage{multirow}
\usepackage{color}
\usepackage{adjustbox}

\usepackage{newfloat}
\usepackage{listings}
\DeclareCaptionStyle{ruled}{labelfont=normalfont,labelsep=colon,strut=off} 
\floatstyle{ruled}
\newfloat{listing}{tb}{lst}{}
\floatname{listing}{Listing}

\usepackage{booktabs}

\title{AngelFingerprint: A Traceable, Explainable, and White-Box Stealthy Watermark for Text-Guided Image Editing}

\author {
    Bo-Han Kung\textsuperscript{\rm 1,\rm 2},
    Futa Waseda\textsuperscript{\rm 2}, 
    Ching-Chun Chang\textsuperscript{\rm 2}, 
    Isao Echizen\textsuperscript{\rm 2}, 
    Shang-Tse Chen\textsuperscript{\rm 1} 
}
\affiliations {
    \textsuperscript{\rm 1}National Taiwan University, Taiwan\\
    \textsuperscript{\rm 2}National Institute of Informatics, Japan\\
    \{d10922019, stchen\}@csie.ntu.edu.tw, \{futa-waseda, ccchang, iechizen\}@nii.ac.jp
}

\begin{document}

\maketitle

\begin{abstract}
Text-guided diffusion editing raises disinformation concerns, making reliable image provenance essential. 
While watermarks are commonly used for this purpose, most methods carry a fixed ID that cannot explain what was changed and which prompt produced it. 
Furthermore, under open-source white-box access, attackers can easily locate and remove watermarks added as separate modules. 
Targeting this setting, we propose AngelFingerprint, a novel watermarking framework ensuring edit traceability, explainability, and white-box stealthiness. It integrates a LoRA into the diffusion model to embed the editing prompt's CLIP text embedding directly into the model's weights. An extractor then recovers this embedding from the image pixels alone. This semantic payload explains the edit, while the weight-integrated design makes it hard to detect and isolate even under full white-box access. Two techniques make this possible: a velocity-alignment anchor that preserves edit quality, and a specially designed frequency filter that keeps the watermark imperceptible yet recoverable and robust. On the MagicBrush dataset, our extractor achieves $86\%$ top-1 accuracy in a 200-way prompt retrieval, versus $20\%$ for prompt inversion.
\end{abstract}

\section{Introduction}
Text-guided diffusion editing has turned image editing into a one-sentence operation \cite{brooks2023instructpix2pix, zhao2024ultraedit}. 
Given an image, any user can produce a realistic edit from a short instruction in seconds. 
In the context of misinformation, such editing can pose a greater risk than generating an image from scratch: because most of the original photograph remains intact, the result retains the credibility of a genuine image.
Manipulations such as fabricated evidence, disinformation, and non-consensual imagery thus become easy and convincing~\cite{chesney2019deepfakes,mirsky2021deepfakes}. To counter this misuse, a common goal is provenance: to reveal whether an image was AI-edited and what was changed.

A natural way to provide provenance is to embed a watermark into the image. It carries an imperceptible payload and can be recovered later to certify origin. 
Most watermarks for generative models carry a fixed payload, such as a model or user ID~\cite{fernandez2023stable,wen2023tree,yang2024gaussian,feng2024aqualora}. 
However, for tracking the misuse of editing, these methods fall short on two fronts.
First, the payload is fixed in advance and says nothing about the edit itself. It can attribute an image to a model, but cannot tell us what was altered or reveal the intent behind the edit. 
Second, the watermark is injected by a separable module, such as an external encoder or a post-hoc module added to the pipeline~\cite{bui2023rosteals,hu2024robustwide}. 
This module is a structurally identifiable part of the system. Because many diffusion models are open-source \cite{rombach2022high, esser2024scaling}, the adversary has full access to the model's architecture, code, and computation.
They can inspect every detail, identify the watermarking module, and remove it. Against such an adversary, any structural or computational difference is a target \cite{feng2024aqualora}.

\begin{figure*}[!t]
  \centering
  \includegraphics[width=1.97\columnwidth]{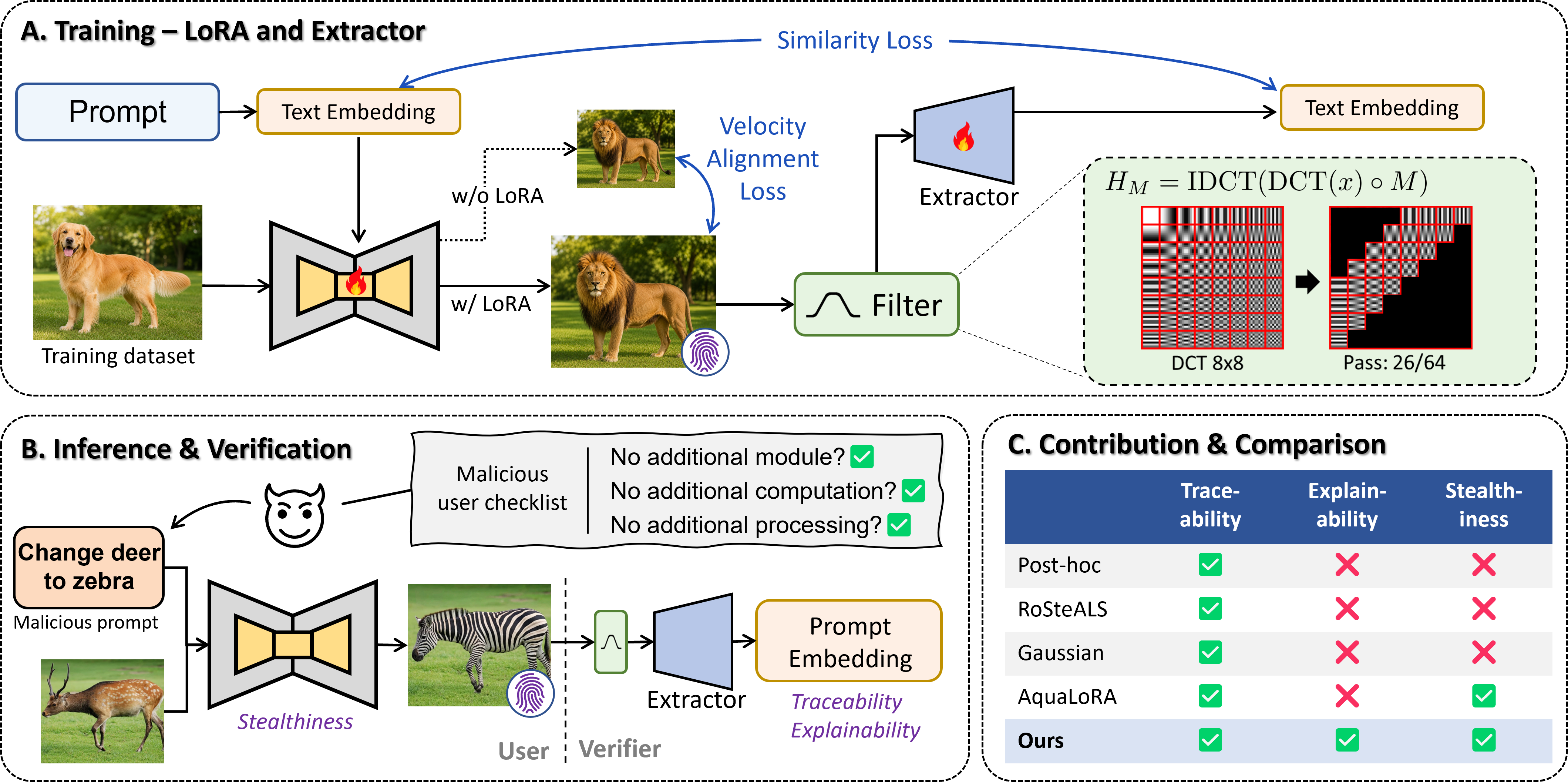}
  \caption{Overview of AngelFingerprint. Part A shows the training pipeline. A LoRA imprints the prompt embedding; then an extractor recovers it. Part B shows the inference scenario: a white-box user can inspect the whole pipeline, find no watermarking module, and run the model without noticing the backdoor. Part C compares AngelFingerprint with prior watermarks.}
  \label{fig:overview} %
\end{figure*}

Therefore, we argue that watermarking AI edits should jointly satisfy three properties: \emph{traceability} (tracing an image's source material), \emph{explainability} (revealing what the edit changed and the possibly malicious intent behind it), and \emph{white-box stealthiness}
(keeping the same architecture and computation as the original, so the watermark is inherently difficult to detect or isolate).
To the best of our knowledge, no prior watermark method achieves all three. 
In this work, we propose a watermark framework that satisfies all.

Concretely, we propose \textbf{AngelFingerprint}. Unlike previous watermarks, it imprints into every edited image an imperceptible watermark of the editing instruction's CLIP text embedding, and a dedicated extractor recovers this embedding from the pixels alone. 
Because the payload is the instruction itself rather than a fixed ID, we can trace the image back to its original prompt and explain what the edit changed.
To achieve stealthiness, the watermark is injected by a low-rank adapter (LoRA)~\cite{hu2022lora}. Thus, only the weight values of UNet/DiT are changed, leaving the architecture, the computation, and the inference code unchanged. A watermarked diffusion model is therefore structurally identical to the original clean model: 
there is no additional module to find, no extra forward pass to notice, and no structural anomalies to detect.
The watermark is a hidden functionality encoded in the model weights: it is silently present, difficult to detect and isolate, and remains hidden from white-box inspection.

Embedding a prompt-based watermark with a LoRA is hard. Unlike a fixed-ID watermark, ours writes a different payload into every edit: the high-dimensional embedding of the edit's prompt. Recovering such a per-sample signal is already demanding, and three further tensions make it worse. 
First, the watermark encodes the prompt, but for some training samples the content already hints at the prompt. The extractor can then ignore the watermark and just guess the prompt from the image content. Such an extractor reads no watermark at all. Training then collapses to image captioning, and the instruction cannot be recovered, which may contain the critical keyword of interest. 
Second, reading and hiding pull against each other. To be read, the watermark must change the image. To stay hidden, it must change the image as little as possible. Pushed too hard, the LoRA distorts the edit and the watermark becomes visible. 
Third, the watermark must survive common image post-processing such as compression and cropping. 
A direct attempt with the AquaLoRA framework \shortcite{feng2024aqualora} confirms the difficulty: training fails, and generated edits are visibly degraded.

We solve these problems with two designs. Both are core technical contributions of AngelFingerprint, and each one is essential. The specially designed \textbf{frequency filter} passes only the middle band of image frequencies to the extractor. This single design solves several problems at once: it stabilizes training, stops the extractor from cheating, and keeps the watermark both invisible and robust. The \textbf{velocity-alignment anchor} keeps the model's editing ability intact, so the edits stay high quality and the watermark stays invisible. Neither is a cosmetic add-on. Without them, training collapses, and no usable watermark emerges. Together, they let a single LoRA carry a hidden, prompt-specific watermark in every edit.


Our contributions can be summarized as follows:
\begin{itemize}
    \item \textbf{Problem.} We formalize watermarking for text-edited images against a white-box, open-weight adversary who controls the whole pipeline. We identify three requirements -- traceability, explainability, and white-box stealthiness -- that prior watermarks satisfy only partially.
    \item \textbf{Method.} We propose AngelFingerprint, a prompt-specific watermark. It imprints the editing instruction's CLIP embedding into images. The watermark satisfies three key requirements at once.
    \item \textbf{Techniques.} We introduce two designs that make this work: a velocity-alignment anchor that preserves edit quality, and a special frequency filter that stabilizes training and keeps the watermark invisible yet robust.
    \item \textbf{Evaluation.} On MagicBrush, our extractor recovers the editing embedding from watermarked images and outperforms prompt-inversion baselines. The watermark also stays imperceptible and robust against post-processing.
\end{itemize}

\begin{table*}[t]
  \centering
  \small
  \setlength{\tabcolsep}{5pt}
  \begin{tabular}{llllcc}
    \toprule
    Method & Watermark carrier & Payload & Capacity & White-box stealthy & Recovers prompt \\
    \midrule
    \multicolumn{6}{l}{\emph{Non-semantic payload watermarks}}\\
    Stable Signature~\shortcite{fernandez2023stable}     & VAE decoder weights      & fixed bit string          & {\raise.17ex\hbox{$\scriptstyle\sim$}}48\,bits & Partial  & No  \\
    Tree-Ring~\shortcite{wen2023tree}                    & initial latent noise     & detection key            & zero-bit         & No  & No  \\
    Gaussian Shading~\shortcite{yang2024gaussian}        & initial latent noise     & fixed bit string          & 256\,bits        & No  & No  \\
    Spherical~\shortcite{hu2026spherical}        & initial latent noise     & fixed bit string          & 512\,bits        & No  & No  \\
    SEAL~\shortcite{arabi2025seal}                       & initial latent noise     & semantics-derived key    & zero-bit         & No  & No  \\
    AquaLoRA~\shortcite{feng2024aqualora}                & LoRA weights (U-Net)     & fixed bit string          & 48\,bits         & Yes & No  \\
    RoSteALS~\shortcite{bui2023rosteals}                 & frozen AE latent         & fixed bit string          & 100\,bits        & No  & No  \\
    Robust-Wide~\shortcite{hu2024robustwide}             & post-hoc encoder         & fixed bit string          & 64\,bits         & No  & No  \\
    \midrule
    \multicolumn{6}{l}{\emph{Semantic payload watermarks}}\\
    SWIFT~\shortcite{evennou2024swift}                   & post-hoc encoder   & caption (coded)          & {\raise.17ex\hbox{$\scriptstyle\sim$}}45\,bits & No  & No  \\
    LatentSeal~\shortcite{evennou2025latentseal}         & post-hoc encoder         & text vector              & 256-d vector      & No  & No  \\
    Of-SemWat~\shortcite{tondi2026semwat}                & post-hoc (DFT mid-band)  & prompt text (coded)      & {\raise.17ex\hbox{$\scriptstyle\sim$}}1--3k\,bits & No & Yes \\
    \textbf{AngelFingerprint (ours)}                                   & LoRA weights & CLIP-L embedding  & 768-d vector & \textbf{Yes} & \textbf{Yes} \\
    \bottomrule
  \end{tabular}
  \\
  \vspace{-2pt}
    \caption{%
    Positioning of AngelFingerprint among watermarks for diffusion models. Methods are grouped by whether their payload carries recoverable semantic content. To the best of our knowledge, ours is the only method that is both white-box stealthy and prompt-recovering. 
    White-box stealthy indicates keeping the same architecture and computation as the original, so the watermark is inherently difficult to detect or isolate.
    }
    \label{tab:positioning}
  \vspace{-10pt}
\end{table*}

\section{Related Work}

\paragraph{Watermarks for generative models.}
Watermarks for diffusion models can be divided into two different categories: post-processing and in-generation.
Table~\ref{tab:positioning} summarizes the methods. 
Most of them carry a non-semantic bit-string payload. 
Stable Signature~\cite{fernandez2023stable} fine-tunes the VAE decoder to carry a fixed bit string. We consider its stealthiness to be partial since the VAE differences are easy to detect and replace \cite{feng2024aqualora}. Tree-Ring~\cite{wen2023tree}, Gaussian Shading~\cite{yang2024gaussian}, and Spherical \cite{hu2026spherical} plant a pattern in the initial noise and extract it back by inverting the diffusion process. SEAL~\cite{arabi2025seal} ties that pattern to image semantics, but still only verifies real or fake. 
RoSteALS~\cite{bui2023rosteals} and Robust-Wide~\cite{hu2024robustwide} add a separate encoder. 
AquaLoRA~\cite{feng2024aqualora} utilizes a U-Net LoRA to inject bit strings during the diffusion process, which satisfies white-box stealthiness. 
However, all of them encode and recover a predefined bit string, not the image-specific editing prompt.

A second line carries a semantic payload. SWIFT~\cite{evennou2024swift} and LatentSeal~\cite{evennou2025latentseal} embed a caption or a text vector, and Of-SemWat~\cite{tondi2026semwat} writes the prompt into the frequency band. The text is chosen by the user, and each method injects the watermark through a separable module that a white-box adversary can locate and remove easily.

\paragraph{Prompt inversion.}
A parallel line recovers the prompt of a generated image without any watermark. Optimization methods such as PEZ~\cite{wen2023hardprompts} and PH2P~\cite{mahajan2024ph2p} search for a prompt whose embedding matches the image, VGD~\cite{kim2025vgd} does so without gradients, and captioners and prompt stealers~\cite{li2023blip2,xiao2024florence2,shen2024promptstealer} read a description straight from images. 
All of them infer the prompt from image content. This is unreliable for text-guided image editing, where the prompt specifies only the intended edit rather than the entire image and the change is often subtle. They also cannot tell an edited image from a real one.
We instead read the instruction prompt from an embedded signal, so recovery reflects the actual edit.

\section{Method}
We first provide an overview of the proposed watermarking framework, \textbf{AngelFingerprint}, which satisfies traceability, explainability, and stealthiness.
Next, we introduce the notation used throughout, followed by the training objective, including the velocity-alignment anchor.
Finally, we introduce a specially designed frequency filter. 

\subsection{System Overview}
Figure~\ref{fig:overview} illustrates our method. Two design choices make it traceable, explainable, and white-box stealthy. We build on Stable Diffusion 3 (SD3)~\cite{esser2024scaling}. 
For white-box stealthiness, we add no external module. We insert a LoRA into the DiT attention projections and embed the watermark directly into the generator's weights, as in AquaLoRA~\shortcite{feng2024aqualora}.
For traceability and explainability, the payload is the editing prompt's CLIP embedding, not a fixed ID. Recovering it lets us trace the image to its instruction and explain what the edit changed.
The LoRA has rank $r{=}96$. Only the LoRA matrices $\theta$ and the extractor $E_\phi$ are trainable; the SD3 backbone, its text encoders, and its VAE stay frozen. 
Given a source image $x$ and an editing instruction $y$ with CLIP-L text embedding $\tau(y)\in\mathbb{R}^{768}$, the LoRA-augmented diffusion model $G_\theta$ produces a watermarked image $G_\theta(x,y)$, from which the extractor recovers $\hat{e}=E_\phi\!\big(H(G_\theta(x,y))\big)$, where $H$ is a frequency filter detailed in Section~\ref{sec:filter}.
Training jointly optimizes $\theta$ and $\phi$ to maximize the similarity between $\hat{e}$ and $\tau(y)$.
We choose to encode the CLIP text embedding rather than a raw text bit string for two main reasons. First, the DiT in SD3 only receives the text embedding, so injecting a raw text bit string would require an extra module and extra computation, which violates stealthiness. 
Second, raw text is discrete and not differentiable, so it would break the joint training of the LoRA and the extractor.
The CLIP embedding is therefore the natural target for the LoRA to embed.


\subsection{Training Objective}

\paragraph{A. Flow-matching loss.}
To preserve SD3's original generative capability, we retain its native rectified-flow training objective when optimizing the LoRA.
For a clean latent $z_0$, noise $\epsilon\sim\mathcal N(0,I)$, timestep $t$, and interpolant $z_t=(1-t)\,z_0+t\,\epsilon$, the target velocity is $v^\star=\epsilon-z_0$. With $v_\theta$ the LoRA-augmented DiT prediction,
\begin{equation}
  \mathcal{L}_{\mathrm{fm}}
  = \mathbb{E}_{z_0,\epsilon,t}\big\lVert v_\theta(z_t,t,\tau(y),x) - v^\star \big\rVert_2^2 .
  \label{eq:fm}
\end{equation}
 
\paragraph{B. Embedding-recovery loss.}
To encode prompt semantics into the generated image as a recoverable watermark, we introduce an embedding-recovery loss.
Over a minibatch of $B$ source-instruction pairs $\{(x_i, y_i)\}_{i=1}^{B}$, the extractor maps each carrier to $\hat e_i = E_\phi\!\big(H(G_\theta(x_i, y_i))\big)$, whose recovery target is the CLIP-L text embedding $\tau_i = \tau(y_i) \in \mathbb{R}^{768}$. For any indices $i, j \in \{1, \dots, B\}$, the cosine similarity between recovered embedding $\hat e_i$ and target $\tau_j$ is $s_{ij} = \cos(\hat e_i, \tau_j) = \langle \hat e_i / \lVert \hat e_i \rVert,\, \tau_j/\lVert \tau_j\rVert\rangle$. The first component is a per-pair alignment loss that pulls each embedding onto its own target,
\begin{equation}
  \mathcal{L}_{\mathrm{cos}} = \frac{1}{B}\sum_{i=1}^{B}\big(1 - s_{ii}\big).
  \label{eq:cos}
\end{equation}
Eq.~\eqref{eq:cos} constrains only the diagonal entries $s_{ii}$ and leaves the off-diagonal similarities $s_{ij}$ ($j\neq i$) free. It can therefore be driven low without making the embedding discriminative: CLIP text embeddings occupy a narrow cone \cite{gao2019representation},
so an extractor that maps every carrier near the shared centroid of the targets already attains high $s_{ii}$, yet $s_{ii}\approx s_{ij}$ and the top-$1$ retrieval $\arg\max_j s_{ij}$ is no better than chance. 
Since verification is a retrieval over a pool, retrieval requires a margin $s_{ii} > s_{ij}$ for every $j\neq i$, which $\mathcal{L}_{\mathrm{cos}}$ does not control. We enforce it with a symmetric in-batch InfoNCE loss~\cite{oord2018cpc}, which treats the $B$ matched pairs as the positives of a $B$-way classification performed in both directions with temperature $\kappa$,
\begin{equation}
  \small
  \mathcal{L}_{\mathrm{nce}} = -\frac{1}{B}\sum_{i=1}^{B}\left[ \log\frac{e^{s_{ii}/\kappa}}{\sum_{j=1}^{B} e^{s_{ij}/\kappa}} + \log\frac{e^{s_{ii}/\kappa}}{\sum_{j=1}^{B} e^{s_{ji}/\kappa}} \right].
  \label{eq:nce}
\end{equation}
The two log-ratios are the image-to-text and text-to-image directions: the first classifies which target $\tau_j$ matches image $i$, the second which recovered embedding $\hat e_j$ matches target $i$, and each denominator sums over the whole batch so that the $B{-}1$ mismatched pairs act as negatives. Minimizing Eq.~\eqref{eq:nce} maximizes $s_{ii}$ relative to the log-sum-exp of its negatives, i.e.\ it directly optimizes the retrieval margin that Eq.~\eqref{eq:cos} leaves unconstrained. The embedding loss is the weighted sum
\begin{equation}
  \mathcal{L}_{\mathrm{embed}} = \lambda_{\mathrm{cos}}\,\mathcal{L}_{\mathrm{cos}} + \lambda_{\mathrm{nce}}\,\mathcal{L}_{\mathrm{nce}}.
  \label{eq:embed}
\end{equation}

\paragraph{C. Velocity-alignment fidelity anchor.}
The embedding recovery loss rewards any change to the carrier that the extractor can read. 
On its own, it lets the LoRA drift the output away from what plain SD3 would produce. This lowers image quality and, worse, makes the watermark visible. We prevent this by anchoring the LoRA-augmented model to the vanilla model in velocity space. 
At deployment, each image is rendered with classifier-free guidance (CFG): for either model, the sampler follows the guided velocity $v_{\mathrm{cfg}}^{m} = v_u^{m} + w\,(v_c^{m} - v_u^{m})$ for $m\in\{\mathrm{lora},\mathrm{van}\}$, where $v_c^{m}$ and $v_u^{m}$ are the predictions with the real instruction and with an empty instruction, $\mathrm{van}$ is vanilla, and $w$ is the guidance scale. 
A naive fix is to align the two guided velocities, $v_{\mathrm{cfg}}^{\mathrm{lora}}$ and $v_{\mathrm{cfg}}^{\mathrm{van}}$.
However, this single constraint holds only at one guidance scale. The two branches can still drift, as long as the drifts cancel at that scale. The guidance scale is a deployment choice, and a user may render at a different one. There the drifts no longer cancel, and the watermarked output separates from the vanilla one.
We therefore align the two branches that build the guided velocity: the conditional pair ($v_c^{\mathrm{lora}}$, $v_c^{\mathrm{van}}$) and the unconditional pair ($v_u^{\mathrm{lora}}$, $v_u^{\mathrm{van}}$). Matching all four predictions keeps the two models equal at every $w$. We call this the four-point alignment. 
Taking $v_c^{\mathrm{van}}$ and $v_u^{\mathrm{van}}$ from a no-grad forward pass of the same DiT without LoRA and treating them as fixed targets (stop-gradient), the anchor is
\begin{equation}
  \mathcal{L}_{\mathrm{ref}} = \alpha\,\big\lVert v_c^{\mathrm{lora}} - \operatorname{sg}(v_c^{\mathrm{van}})\big\rVert_2^2 + \beta\,\big\lVert v_u^{\mathrm{lora}} - \operatorname{sg}(v_u^{\mathrm{van}})\big\rVert_2^2 ,
  \label{eq:ref}
\end{equation}
where $\operatorname{sg}(\cdot)$ is the stop-gradient. 
We weight the unconditional branch more heavily ($\beta > \alpha$): it is shared across all prompts and forms the base that guidance extrapolates from, so its drift is the most damaging to fidelity, whereas the conditional branch is anchored more loosely to leave the LoRA the capacity it needs to write a readable watermark.

\paragraph{D. Full objective.}
The joint objective over $(\theta,\phi)$ is
\begin{equation}
  \mathcal{L}
  = \lambda_{\mathrm{fm}}\,\mathcal{L}_{\mathrm{fm}}
  + \lambda_{\mathrm{embed}}\,\mathcal{L}_{\mathrm{embed}}
  + \lambda_{\mathrm{ref}}\,\mathcal{L}_{\mathrm{ref}} .
  \label{eq:total}
\end{equation}
For our main model we set
$\lambda_{\mathrm{fm}}{=}0.5$, $\lambda_{\mathrm{embed}}{=}2.0$
(with $\lambda_{\mathrm{cos}}{=}1.0$, $\lambda_{\mathrm{nce}}{=}1.0$,
$\kappa{=}0.07$), and $\lambda_{\mathrm{ref}}{=}3.0$ with
$(\alpha,\beta){=}(1.0,2.0)$. We provide the ablation studies in Appendix B.

\begin{table*}[t]
  \centering
  \small
  \setlength{\tabcolsep}{5pt}
  \begin{tabular}{lccccccccc}
    \toprule
    & \multicolumn{3}{c}{Embedding similarity} & \multicolumn{6}{c}{Retrieval} \\
    \cmidrule(lr){2-4}\cmidrule(lr){5-10}
    Method & $\cos_{\mathrm{p}}\uparrow$ & $\cos_{\mathrm{off}}$ & $\Delta_{\cos}\uparrow$ & Top-1$\uparrow$ & Top-1$_{\mathrm{off}}$ & $\Delta_{\mathrm{T1}}\uparrow$ & Top-5$\uparrow$ & Top-10$\uparrow$ & MRR$\uparrow$ \\
    \midrule
    BLIP-2~\cite{li2023blip2}             & 0.550 & 0.430 & 0.120 & 0.215 & 0.330 & $-0.115$ & 0.510 & 0.650 & 0.354 \\
    PromptStealer~\cite{shen2024promptstealer} & 0.099 & 0.005 & 0.094 & 0.215 & 0.350 & $-0.135$ & 0.495 & 0.570 & 0.342 \\
    Florence-2~\cite{xiao2024florence2}   & 0.158 & 0.026 & 0.132 & 0.225 & 0.315 & $-0.090$ & 0.505 & 0.660 & 0.363 \\
    VGD~\cite{kim2025vgd}                 & 0.385 & 0.301 & 0.084 & 0.160 & 0.305 & $-0.145$ & 0.405 & 0.490 & 0.277 \\
    \midrule
    \textbf{Ours (UltraEdit)}      & {0.750} & 0.378 & {0.372} & {0.655} & 0.005 & ${+0.650}$ & {0.950} & {0.970} & {0.780} \\
    \textbf{Ours (SD3 medium)}             & \textbf{0.802} & 0.364 & \textbf{0.438} & \textbf{0.860} & 0.000 & \textbf{$+$0.860} & {0.990} & {1.000} & \textbf{0.917} \\
    \bottomrule
  \end{tabular}
  \vspace{-2pt}
    \caption{%
    Prompt recovery from the watermarked image, and the watermark's contribution to it. The subscript ``$\mathrm{off}$'' marks the value on the unwatermarked image (LoRA disabled). $\cos_{\mathrm{p}}$ is the cosine between the recovered embedding and the true prompt embedding. $\Delta_{\cos}{=}\cos_{\mathrm{p}}{-}\cos_{\mathrm{off}}$ and $\Delta_{\mathrm{T1}}{=}$Top-1$-$Top-1$_{\mathrm{off}}$ are how much the watermark adds to the cosine and to Top-1. 
    Retrieval is over a pool of $N{=}200$ MagicBrush prompts.}
  \label{tab:main}
\end{table*}

\subsection{Frequency Filtering}
\label{sec:filter}
Jointly training the LoRA and the extractor from the objective above is unstable and difficult to optimize. The two networks co-adapt, and without an additional mechanism, the LoRA cannot render high-quality watermarked images. 
Our key design is a frequency filter $H$, a fixed band-pass filter inserted between the generated image and the extractor that passes only a mid-frequency band. 
It makes the joint optimization tractable and, at the same time, fixes where the watermark lives in the spectrum. 
The filter is essential rather than cosmetic. Without it, training still makes some progress, but never reaches a usable model: the carrier becomes visibly degraded, and the recovered prompt is far weaker.

Concretely, $H$ transforms the watermarked image $G_\theta(x,y)$ into the frequency domain and applies a fixed, predefined mask $M$ that keeps a mid-frequency band and zeros both the low- and high-frequency coefficients before returning the masked signal to the extractor. 
The blocked coefficients are dropped in the forward pass, so they also receive no gradient: the extractor sees only the mid-band, and no embedding-recovery gradient reaches the rest of the spectrum. 
The mask is set once, before training, and it never changes during training or inference.

This band-pass filtering is central to AngelFingerprint. It brings four benefits:
  \paragraph{Convergence.} 
 Most image semantics lie in the low-frequency band. Without filtering, the extractor predominantly drives the LoRA to embed the watermark in this band. However, the flow-matching and fidelity losses strive to keep this band completely unchanged. These conflicting gradients destabilize training. By filtering out the low-frequency band from the extractor, we force the watermark into the mid-frequency band, leading to more stable optimization.
      \paragraph{Preventing semantic shortcut.} 
  Because the target edit-prompt embedding correlates with image content, the extractor can reduce the embedding-recovery loss by exploiting image semantics rather than decoding the embedded watermark. Blocking the content-dominant frequency band removes this shortcut and forces the extractor to rely on the watermark signal.
  \paragraph{Imperceptibility.} The low band is zeroed before the extractor, so the extractor never reads a low-frequency signal. The LoRA gains nothing by writing there, so it leaves the low frequencies alone. These frequencies dominate the image's appearance, so the edit looks unchanged. The watermark simply cannot live in the visible band. 
  \paragraph{Robustness.} The highest frequencies are the first to be lost to common post-processing, such as compression and resampling. Excluding them keeps the watermark in the more stable mid-band. The watermark signal then survives these operations.


To instantiate this filter, we design a discrete cosine transform (DCT) band-pass mask that explicitly mirrors the JPEG algorithm, the most ubiquitous image compression standard. JPEG transforms non-overlapping $8 \times 8$ blocks into the DCT domain and discards high-frequency coefficients. By restricting our watermark exclusively to the $26/64$ mid-frequency coefficients (Figure~\ref{fig:overview}), we avoid both the low-frequency band (to stabilize training, improve fidelity, and prevent semantic shortcuts) and the high-frequency band (to ensure robustness against compression). Consequently, the watermark is robust to medium- and high-quality JPEG compression by construction. Mask design details are deferred to Appendix B.

Our two techniques, the frequency filter and the velocity-alignment anchor, are complementary rather than competing. The filter controls where the watermark lives: it admits the mid-band and forbids the visible low frequencies. The anchor controls how far the output may drift from vanilla SD3. The two act on independent levers. The visible low frequencies are thus protected twice: the filter keeps the watermark out, and the anchor pins them to the frozen model. Imperceptibility and recoverability then improve together instead of trading off.


\newcommand{\na}{\ensuremath{\text{--}}}
\newcommand{\nr}{\textit{\scriptsize n/r}}

\begin{table*}[t]
  \centering
  \small
  \setlength{\tabcolsep}{4pt}
  \begin{tabular}{lcccccccc}
    \toprule
    & \multicolumn{6}{c}{Visual fidelity (LoRA on vs.\ off)} & Retrieval \\
    \cmidrule(lr){2-7}\cmidrule(lr){8-8}
    Method / extractor-input filter & DreamSim$\downarrow$ & LPIPS$\downarrow$ & PSNR$\uparrow$ & SSIM$\uparrow$ & FID$\downarrow$ & KID$\downarrow$ & Top-1$\uparrow$ \\
    \midrule
     w/o any filter & 0.319\,\tiny$\pm$0.092 & 0.366\,\tiny$\pm$0.060 & 20.85\,\tiny$\pm$1.64 & 0.650\,\tiny$\pm$0.095 & 156.2 & 0.0093\,\tiny$\pm$0.0053 & 0.605 \\
    Gaussian high-pass ($\sigma{=}2$) & 0.222\,\tiny$\pm$0.091 & \textbf{0.297}\,\tiny$\pm$0.062 & \textbf{22.92}\,\tiny$\pm$2.12 & 0.704\,\tiny$\pm$0.097 & 133.6 & 0.0085\,\tiny$\pm$0.0037 & 0.805 \\
    \textbf{JPEG mid-band mask (ours)} & \textbf{0.221}\,\tiny$\pm$0.083$^\dagger$ & 0.299\,\tiny$\pm$0.062 & 22.91\,\tiny$\pm$2.02 & \textbf{0.704}\,\tiny$\pm$0.095 & \textbf{132.1} & \textbf{0.0032}\,\tiny$\pm$0.0029 & \textbf{0.860} \\
    \bottomrule
    \multicolumn{8}{l}{$^\dagger$ The DreamSim values of TreeRing~\shortcite{wen2023tree} and AquaLoRA~\shortcite{feng2024aqualora} {\tiny(SD1.5, 48-bit)} are 0.301 and 0.201, respectively} 
  \end{tabular}
  \vspace{-8pt}
    \caption{%
    This table demonstrates the imperceptibility of the watermark. We measure the visual similarity between the watermarked image (LoRA on) and the original output image (LoRA off) using different metrics. 
    }
  \label{tab:visual}
  \vspace{-6pt}
\end{table*}

\section{Experiments}
We evaluate AngelFingerprint on different aspects: watermark performance, white-box stealthiness, visual fidelity, and robustness, while also conducting ablation studies. Experiments are performed on the MagicBrush benchmark~\cite{zhang2023magicbrush} using SD3-medium and UltraEdit~\cite{zhao2024ultraedit} as backbone models. Full model specifications, dataset configurations, and implementation details are provided in Appendix D.

\subsection{Watermark Performance}
Our watermark embeds the editing prompt's embedding into each edit and recovers it after generation. We therefore evaluate two things: how close the recovered embedding is to the true prompt embedding, and how well it retrieves the correct prompt from a $200$ candidate prompt pool. 

We first report $\cos_{\mathrm{p}}$, the cosine similarity between the recovered embedding and the true prompt embedding. To check that this similarity comes from the watermark, we also measure $\cos_{\mathrm{off}}$, the same cosine on the unwatermarked image, and report the gain $\Delta_{\cos}{=}\cos_{\mathrm{p}}{-}\cos_{\mathrm{off}}$. A large gain means the watermark, not the image content, drives the recovery. This is exactly what our filter enforces: it removes the image content from the extractor's input, so the extractor cannot recover the prompt by reading the image and is forced to read the embedded watermark instead.
We then retrieve the recovered embedding against a pool of $N{=}200$ prompts from the MagicBrush test set. We report Top-$k$ accuracy and mean reciprocal rank (MRR)~\cite{craswell2009mean}. This measures whether we can identify the exact editing instruction within a pool via the recovered embedding.
Since our payload is a prompt embedding, fixed-bit watermarking methods are not directly comparable. We therefore compare against captioning and prompt-inversion baselines, which are the relevant alternatives for recovering an editing instruction from an image under the white-box setting.


Table~\ref{tab:main} reports prompt recovery on the watermarked image. Our method recovers the editing instruction far better than every baseline. On SD3 medium, the recovered embedding is close to the true prompt ($\cos_{\mathrm{p}}{=}0.802$), and retrieval reaches Top-1 $0.860$ and MRR $0.917$. The advantage also holds on UltraEdit~\cite{zhao2024ultraedit} (Top-1 $0.655$), so it is not tied to one backbone.

This gain comes from the watermark, not the image content. We run each method on the watermarked and the unwatermarked image (subscript ``off'') and report the gains $\Delta_{\cos}$ and $\Delta_{\mathrm{T1}}$. For our extractor both are large ($\Delta_{\cos}{=}0.438$, $\Delta_{\mathrm{T1}}{=}{+}0.860$), so the watermark drives recovery. Without it, retrieval drops to chance. The baselines instead read the content either way, so their $\Delta_{\cos}$ is small and their $\Delta_{\mathrm{T1}}$ is negative. Our extractor therefore reads the watermark alone.

\begin{figure}
  \centering
  \includegraphics[width=\linewidth]{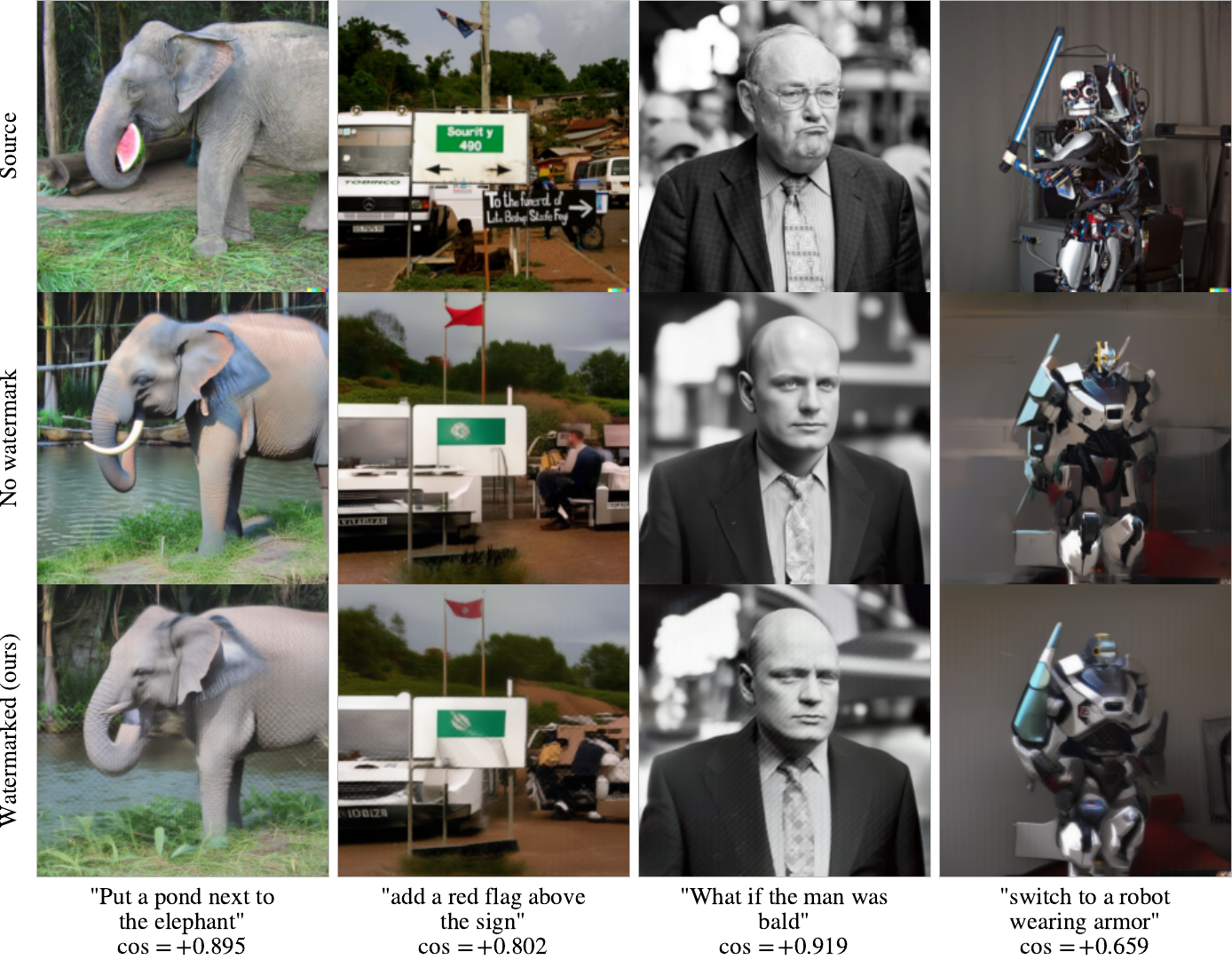}
  \vspace{-12pt}
  \caption{%
    Qualitative watermarked edits under deployment sampling. The watermark leaves only a faint, grid-like texture, hard to spot without close inspection. }
  \label{fig:real_world}
\end{figure}

\subsection{White-box Stealthiness and Visual Fidelity}
To survive a white-box adversary, the watermark must hide at two levels: in the model, and in the images it produces. AngelFingerprint changes no code, no computation, and no architecture of SD3. It only changes weight values. The watermarked model is therefore structurally identical to an ordinary one, and a provider can release it as an ordinary fine-tuned model. There is no module to isolate, so the watermark is hard to locate and hard to remove, like the in-weights watermark AquaLoRA~\cite{feng2024aqualora}.

In addition to model stealthiness, the watermark should be imperceptible in the output. Table~\ref{tab:visual} compares the images generated with and without the watermark. Note that the watermark is injected during generation, so a watermarked image has no ground-truth clean image without watermark. We use the vanilla SD3 output (LoRA off) as the reference. Pixel metrics like PSNR and SSIM are therefore only loosely meaningful, and AquaLoRA omits them for this reason; we still report them to show the effect of the filter. Across every metric the watermarked image stays close to the LoRA-off output. Figure~\ref{fig:real_world} shows qualitative examples: the watermark appears as a faint, grid-like texture that is hard to spot without close inspection.

We also check realism. Table~\ref{tab:realism} reports FID and KID against real MagicBrush images. The first row, vanilla SD3, is the realism floor of the clean model. With the JPEG mid-band filter our FID and KID sit close to this floor, so the watermark barely shifts the output distribution away from clean SD3.

\begin{table}[t]
  \centering \small
  \setlength{\tabcolsep}{6pt}
  \begin{tabular}{lcc}
    \toprule
    & \multicolumn{1}{c}{FID $\downarrow$} & \multicolumn{1}{c}{KID $\downarrow$ ($\times10^{-3}$)} \\
    \midrule
    Vanilla SD3 w/o LoRA & 158.6 & 8.47 \tiny$\pm$ 2.79  \\
    \midrule
    No filter &  181.3  & 26.57 \tiny$\pm$ 5.07 \\
    Gaussian high-pass & 173.1  & 24.59 \tiny$\pm$ 3.98 \\
    \textbf{JPEG mid-band mask}  & \textbf{164.8}  & \textbf{13.62 \tiny$\pm$ 3.18} \\
    \bottomrule
  \end{tabular}
  \vspace{-6pt}
    \caption{%
  We measure the visual fidelity of the generated images with respect to real images in the MagicBrush target set. The baseline is the vanilla SD3 without LoRA. It illustrates the importance of the filter, where it makes the generated images of LoRA more similar to the original.
  }
  \label{tab:realism}
\end{table}

\begin{table}[t]
  \centering
   \begin{adjustbox}{max width=\columnwidth}
  \begin{tabular}{lccccc}
        \toprule
    & \multicolumn{3}{c}{Watermark survival} & \multicolumn{2}{c}{Generation quality} \\
    \cmidrule(lr){2-4}\cmidrule(lr){5-6}
    & Top-1$\uparrow$ & MRR$\uparrow$ & $\cos_{\mathrm{p}}\uparrow$ & KID$\downarrow$ ({\tiny$\times10^{-3}$}) & DreamSim$\downarrow$ \\
    \midrule
    Baseline                & \textbf{0.860} & 0.917 & +0.802 & 14.3 & 0.221 \\
    \midrule
    \multicolumn{6}{l}{\emph{Continued fine-tuning (steps)}}\\
    \quad 50                        & 0.840 & 0.905 & +0.779 & 10.2 & 0.229 \\
    \quad 100                       & 0.750 & 0.843 & +0.737 & 11.0 & 0.230 \\
    \quad 200                       & 0.650 & 0.760 & +0.686 & 11.5 & 0.244 \\
    \quad 300                       & 0.575 & 0.699 & +0.653 & 10.1 & 0.248 \\
    \midrule
    \multicolumn{6}{l}{\emph{Weight quantization (precision)}}\\
    \quad fp16                      & 0.860 & 0.917 & +0.802 & 13.4 & 0.222 \\
    \quad bf16                      & 0.855 & 0.915 & +0.802 & 14.3 & 0.221 \\
    \quad int8                      & 0.860 & 0.918 & +0.803 & 13.0 & 0.218 \\
    \quad int4                      & 0.505 & 0.648 & +0.685 & 164.4 & 0.088 \\
    \bottomrule
  \end{tabular}
\end{adjustbox}
  \vspace{-6pt}
  \caption{%
  Robustness of AngelFingerprint to two operations that a user/attacker can apply after release: fine-tuning and low-precision quantization of the released weights.
    }
  \label{tab:attack}
\end{table}

\subsection{Robustness}
\begin{figure*}[t]
  \centering
  \includegraphics[width=\textwidth]{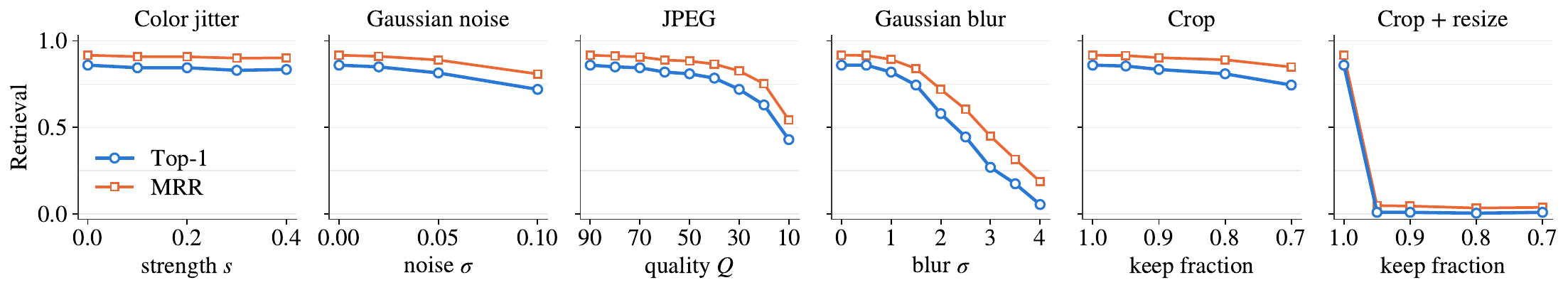}
  \vspace{-19pt}
  \caption{%
   Robustness of the watermark under post-processing attacks. The proposed watermark is robust against most post-processing attacks. If the post-processing operation does not destroy the mid-frequency band, the watermark survives. Resizing is the main threat to the proposed design, because it resamples the frequency components. Nevertheless, the watermark is robust to mild photometric attacks, such as color jitter, Gaussian noise, and JPEG compression.}
  \label{fig:robustness}
\end{figure*}

\begin{figure*}[t]
  \centering
  \includegraphics[width=0.95\textwidth]{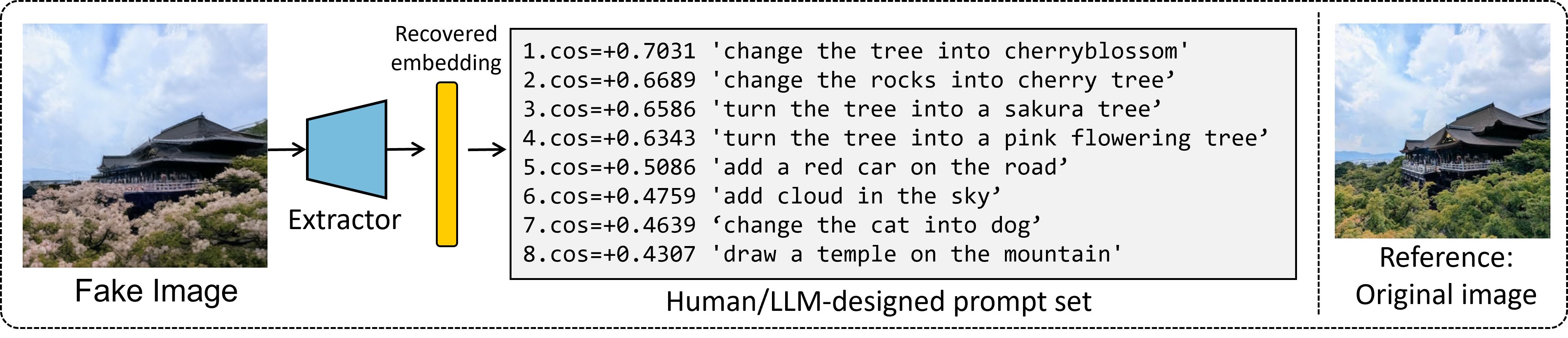}
  \vspace{-10pt}
  \caption{%
    Open-set verification on a real photograph. We edit it with AngelFingerprint using the prompt ``change the tree into cherry blossom,'' then recover the embedding and score candidate sentences against it by cosine similarity. Only the true prompt and semantically similar sentences score high, while unrelated sentences stay low. The recovered embedding is thus semantically meaningful, so a verifier can identify the edit without knowing the exact wording. }
  \label{fig:real_world_demo}
\end{figure*}

Since watermarked images may be modified when shared online, we evaluate the robustness of the watermark against common post-processing attacks, including color jitter, Gaussian noise, JPEG compression, Gaussian blur, cropping, and resizing. The results are shown in Figure~\ref{fig:robustness}. As expected, the watermark survives whenever the post-processing leaves the mid-frequency band intact.
In contrast, if the operation destroys the mid-frequency band, the watermark is lost. Resizing is the clearest example: it desynchronizes the $8\times 8$ DCT blocks and scrambles the mid-frequency band.
In practice, the most common operations are JPEG compression and cropping, and the watermark survives both.

A white-box adversary holds the released weights and can tune them directly. We therefore test two weight-level attacks, continued fine-tuning and low-precision quantization (Table~\ref{tab:attack}). Quantization barely affects the watermark: down to int8, Top-1 stays at $0.86$ with no loss in quality. Only int4 lowers it (Top-1 $0.505$), but int4 also breaks the model, with KID rising from $14.3$ to $164.4$, so the attacker is left with an unusable editor. Fine-tuning degrades the watermark only gradually. After $50$ steps, Top-1 is still $0.840$, and even after $300$ steps it holds at $0.575$, far above chance. Removing the watermark this way costs real compute and never fully succeeds.

\subsection{Ablation Study}
We conduct several ablation studies, including each term in our main objective function, the hyperparameters in the loss function, and different band-pass filter configurations. All of them can be found in Appendix B.

\subsection{Real-world Deployment}
The benchmark above retrieves against a fixed pool of $200$ prompts in the dataset. A real verifier may have no such pool. We therefore test an open-set setting. We take a real photograph and use our watermarked model to produce an AI-generated image. The verifier then recovers the embedding from the edited image alone via the extractor.
The verifier builds their own candidate set. They can write sentences that match their suspicion, or generate many candidates with a vision-language model. Each candidate is scored by cosine similarity to the recovered embedding, and the closest ones are returned.
Because the payload is a semantic embedding, the exact wording is not needed. Figure~\ref{fig:real_world_demo} edits a photograph. The ground truth sentence and its close paraphrases score high, and unrelated sentences score low. The verifier can therefore tell what the edit changed, and the intent behind it. More results are in Appendix C.

\section{Conclusion}

We presented AngelFingerprint, a watermarking framework that unifies traceability, explainability, and white-box stealthiness for text-guided diffusion editing. By embedding the prompt's CLIP embedding into LoRA weights attached to the generator, our approach avoids a separable watermarking module. Our results demonstrate robust prompt recovery that substantially outperforms standard inversion baselines. Ultimately, this work points toward accountable open-weight releases: providers can distribute models that inherently embed a self-explaining record into every generated edit. A natural next step is to extend the extractor to decode these payload embeddings directly into text.



\clearpage

\bibliography{aaai2027}

\appendix

\setcounter{table}{0}
\renewcommand{\thetable}{A\arabic{table}}
\setcounter{figure}{0}
\renewcommand{\thefigure}{A\arabic{figure}}

\section{More Experimental Results}

\subsection{Watermark Analysis}
To measure the watermark's contribution to prompt retrieval, we compare Top-1 accuracy and MRR on watermarked and unwatermarked images. Table~\ref{tab:contribution} reports the results. By design, the filter prevents the extractor from relying on original image content. The near-zero retrieval accuracy on unwatermarked images shows that the extractor is not a trivial CLIP image encoder; instead, it reads the signal embedded by AngelFingerprint.

\subsection{Robustness}
\label{sec:robustness}
We evaluate robustness from two perspectives: robustness to image post-processing and robustness to post-hoc fine-tuning.
\subsubsection{Robustness Against Post-Processing}
Table~\ref{tab:robustness} provides the detailed post-processing results. Because the filter constrains the watermark to the DCT mid-band, the watermark survives distortions that preserve this band. When a distortion disrupts the mid-band spectrum, recovery can fail. For example, resizing resamples the image and desynchronizes the $8{\times}8$ DCT blocks, making AngelFingerprint less robust to crop-and-resize operations. In contrast, robustness to JPEG compression is expected by construction because the DCT-based filter places the watermark in coefficients that medium- and high-quality JPEG compression tends to preserve. Cropping and compression are also among the most common post-processing operations for images shared online, making robustness to these operations especially important in real-world deployment. As Table~\ref{tab:robustness} shows, AngelFingerprint remains robust to both cropping and JPEG compression.

\subsubsection{Robustness Against Post-Hoc Fine-Tuning}
The watermark lives in a LoRA on the diffusion backbone, so in practice a released model ships as the base Stable Diffusion~3 weights merged with this adapter. We study a strong white-box case for such a release, in which a downstream user continues to fine-tune the released model by training a fresh LoRA on top of it. The adversary can inspect all weights but does not know that the watermark exists and never accesses our extractor; their goal is simply to adapt the model to a small set of personal images, as in ordinary LoRA personalization.

The attacker starts from the proposed watermarked model and fine-tunes an additional LoRA while keeping the base transformer, VAE, text encoders, and our extractor frozen. Training uses the standard rectified-flow objective of Stable Diffusion~3, $\lVert v_\theta(z_t,t,x,y)-(\epsilon-z_0)\rVert_2^2$. We use 20 images from MagicBrush (excluded from both the training set and the 200-image test set), AdamW with learning rate $10^{-4}$ and batch size 2, and train for up to 300 steps, saving checkpoints at 0, 50, 100, 200, and 300 steps. The step-0 checkpoint is the untouched released model and reproduces the original watermark exactly.
At each checkpoint, we use it to edit the images in the test set and run our extractor without modification. We score the watermark using top-$k$ retrieval, MRR, and cosine similarity against the true CLIP-L prompt embeddings. In parallel, we measure the attack's cost to generation quality using KID and DreamSim. Table 5 in the main paper reports the results.

\begin{figure}[t]
  \includegraphics[width=\columnwidth]{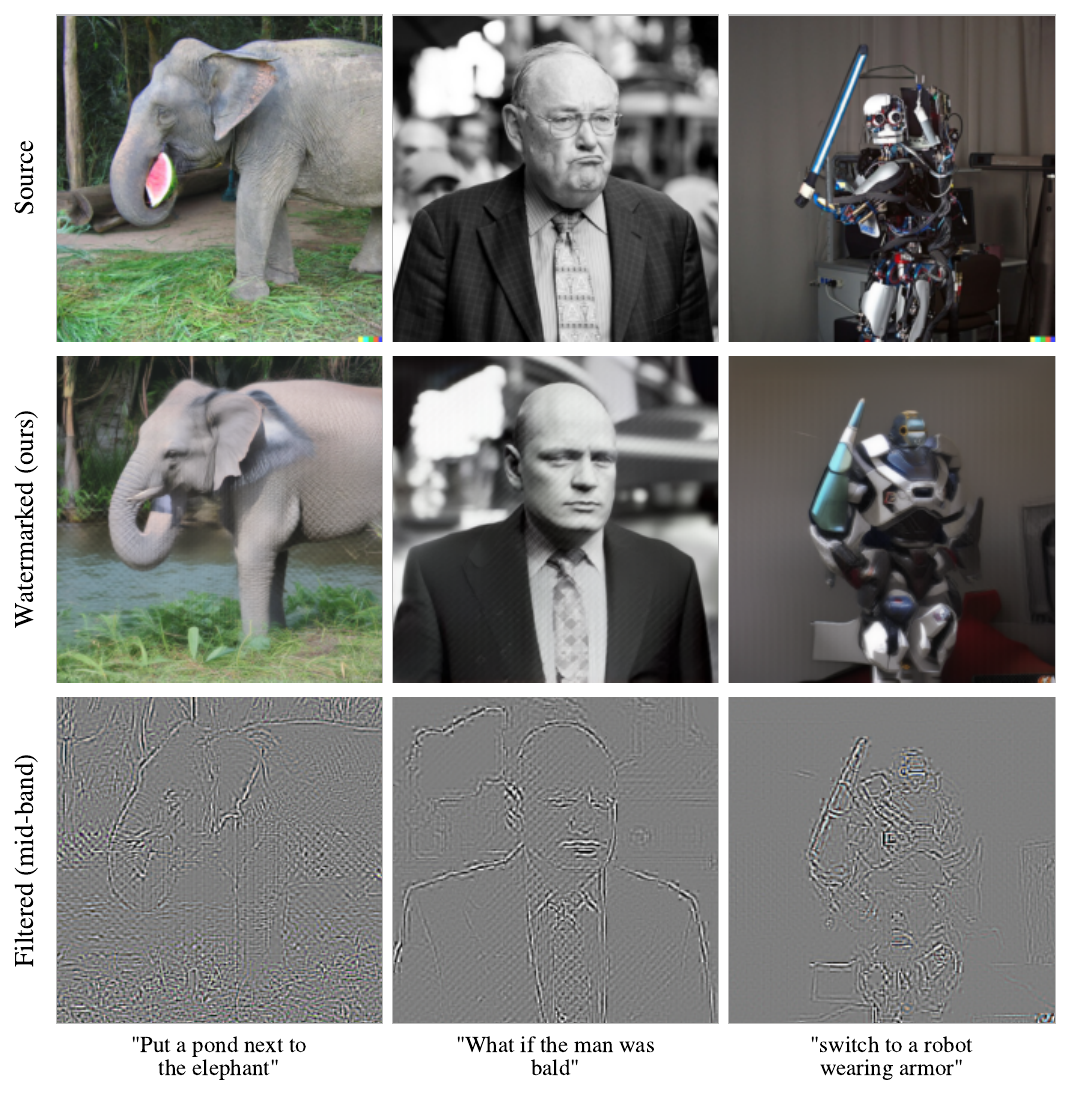}
  \caption{%
  Visualization of the band-pass filtered signal. The first row shows the source image, the second row shows the watermarked image before filtering, and the third row shows the filtered image. The filter removes most semantic content, preventing the extractor from using image content as a shortcut. }
  \label{fig:mid_band_vis}
\end{figure}

\subsection{Watermark Visualization}
Figure~\ref{fig:mid_band_vis} visualizes the isolated watermark signal. Notably, this visualization is not a simple pixel-wise difference between the watermarked and unwatermarked images; rather, it represents the output of our JPEG-based mid-band filter. 
Note that the filtered images are also the input of the extractor.
This filter restricts the frequency spectrum available to the LoRA and the extractor, effectively preventing the extractor from exploiting semantic shortcuts. As observed, the resulting mid-band signal primarily consists of the contours and textures of the generated image.

\begin{table*}[ht]
  \centering
  \setlength{\tabcolsep}{6pt}
  \caption{%
    Isolating the watermark channel. ``no WM'' scores each method on the unwatermarked image (LoRA disabled); ``WM'' scores it on the deployed watermarked image. $\Delta$ is the gain the watermark provides to that reader. Only our extractor can read the injected watermark signal ($\Delta{=}{+}0.860$); the baseline methods cannot exploit it ($\Delta{<}0$), confirming the watermark is a private signal and is invisible to content-based recovery. Our extractor also scores near-zero on the unwatermarked image, i.e.\ it reads the injected signal rather than image content.}
  \label{tab:contribution}
  \begin{tabular}{lcccccc}
    \toprule
    & \multicolumn{3}{c}{Top-1} & \multicolumn{3}{c}{MRR} \\
    \cmidrule(lr){2-4}\cmidrule(lr){5-7}
    Method & no WM & WM & $\Delta$ & no WM & WM & $\Delta$ \\
    \midrule
    BLIP-2~\cite{li2023blip2}             & 0.330 & 0.215 & $-0.115$ & 0.455 & 0.354 & $-0.101$ \\
    PromptStealer~\cite{shen2024promptstealer} & 0.350 & 0.215 & $-0.135$ & 0.464 & 0.342 & $-0.122$ \\
    Florence-2~\cite{xiao2024florence2}   & 0.315 & 0.225 & $-0.090$ & 0.444 & 0.363 & $-0.081$ \\
    VGD~\cite{kim2025vgd}                 & 0.305 & 0.160 & $-0.145$ & 0.409 & 0.277 & $-0.132$ \\
    \midrule
    \textbf{Ours (extractor)}             & 0.000 & \textbf{0.860} & $\mathbf{+0.860}$ & 0.034 & \textbf{0.917} & $\mathbf{+0.883}$ \\
    \bottomrule
  \end{tabular}
\end{table*}

\begin{table*}[t]
  \centering \small
  \setlength{\tabcolsep}{5pt}
  \caption{%
    Post-processing robustness of the watermark. $\cos_{\mathrm{p}}$ is the cosine similarity between recovered embedding and the true prompt embedding, and $\cos_{\mathrm{gap}}$ is its separation margin. \textbf{Ret$_1$} and \textbf{Ret$_{\mathrm{gap}}$} denote Top-1 and margin retention relative to the clean model (\%). Cropping and JPEG compression are the most common operations on shared images, and the watermark survives both. It is fragile only to resizing, which desynchronizes the DCT band.}
  \label{tab:robustness}
  \begin{tabular}{llcccccccc}
    \toprule
    & & \multicolumn{4}{c}{Retrieval} & \multicolumn{2}{c}{Embedding} & \multicolumn{2}{c}{Retention\,\%} \\
    \cmidrule(lr){3-6}\cmidrule(lr){7-8}\cmidrule(lr){9-10}
    Attack & Level & Top-1 & Top-5 & Top-10 & MRR & $\cos_{\mathrm{p}}$ & $\cos_{\mathrm{gap}}$ & Ret$_1\uparrow$ & Ret$_{\mathrm{gap}}\uparrow$ \\
    \midrule
    Clean & -- & 0.860 & 0.990 & 1.000 & 0.917 & +0.802 & +0.438 & 100 & 100 \\
    \midrule
    \multirow{4}{*}{Color jitter} & $s{=}0.1$ & 0.845 & 0.985 & 0.995 & 0.909 & +0.800 & +0.435 & 98 & 99 \\
     & $s{=}0.2$ & 0.845 & 0.985 & 0.995 & 0.909 & +0.798 & +0.433 & 98 & 99 \\
     & $s{=}0.3$ & 0.830 & 0.980 & 0.990 & 0.900 & +0.796 & +0.431 & 97 & 98 \\
     & $s{=}0.4$ & 0.835 & 0.980 & 0.990 & 0.902 & +0.794 & +0.429 & 97 & 98 \\
    \midrule
    \multirow{3}{*}{Gaussian noise} & $\sigma{=}0.02$ & 0.850 & 0.985 & 0.995 & 0.911 & +0.801 & +0.437 & 99 & 100 \\
     & $\sigma{=}0.05$ & 0.815 & 0.980 & 0.995 & 0.890 & +0.790 & +0.421 & 95 & 96 \\
     & $\sigma{=}0.1$ & 0.720 & 0.935 & 0.955 & 0.809 & +0.725 & +0.348 & 84 & 80 \\
    \midrule
    \multirow{9}{*}{JPEG} & $Q{=}90$ & 0.860 & 0.990 & 0.995 & 0.917 & +0.801 & +0.436 & 100 & 100 \\
     & $Q{=}80$ & 0.850 & 0.990 & 1.000 & 0.913 & +0.800 & +0.434 & 99 & 99 \\
     & $Q{=}70$ & 0.845 & 0.985 & 0.995 & 0.908 & +0.795 & +0.428 & 98 & 98 \\
     & $Q{=}60$ & 0.820 & 0.980 & 0.990 & 0.890 & +0.791 & +0.422 & 95 & 97 \\
     & $Q{=}50$ & 0.810 & 0.980 & 0.985 & 0.884 & +0.786 & +0.417 & 94 & 95 \\
     & $Q{=}40$ & 0.785 & 0.970 & 0.985 & 0.866 & +0.775 & +0.405 & 91 & 93 \\
     & $Q{=}30$ & 0.720 & 0.960 & 0.985 & 0.827 & +0.757 & +0.384 & 84 & 88 \\
     & $Q{=}20$ & 0.630 & 0.900 & 0.925 & 0.752 & +0.722 & +0.344 & 73 & 79 \\
     & $Q{=}10$ & 0.430 & 0.670 & 0.750 & 0.544 & +0.632 & +0.240 & 50 & 55 \\
    \midrule
    \multirow{8}{*}{Gaussian blur} & $\sigma{=}0.5$ & 0.860 & 0.985 & 0.995 & 0.916 & +0.800 & +0.435 & 100 & 99 \\
     & $\sigma{=}1$ & 0.820 & 0.975 & 0.990 & 0.893 & +0.788 & +0.418 & 95 & 95 \\
     & $\sigma{=}1.5$ & 0.745 & 0.955 & 0.985 & 0.840 & +0.751 & +0.374 & 87 & 85 \\
     & $\sigma{=}2$ & 0.580 & 0.900 & 0.950 & 0.720 & +0.698 & +0.317 & 67 & 73 \\
     & $\sigma{=}2.5$ & 0.445 & 0.820 & 0.915 & 0.604 & +0.650 & +0.272 & 52 & 62 \\
     & $\sigma{=}3$ & 0.270 & 0.685 & 0.825 & 0.450 & +0.598 & +0.226 & 31 & 52 \\
     & $\sigma{=}3.5$ & 0.175 & 0.460 & 0.635 & 0.316 & +0.542 & +0.175 & 20 & 40 \\
     & $\sigma{=}4$ & 0.055 & 0.340 & 0.450 & 0.188 & +0.486 & +0.116 & 6 & 26 \\
    \midrule
    \multirow{4}{*}{Crop (no resize)} & keep\,$0.95$ & 0.855 & 0.990 & 1.000 & 0.915 & +0.801 & +0.438 & 99 & 100 \\
     & keep\,$0.9$ & 0.835 & 0.985 & 0.995 & 0.903 & +0.800 & +0.433 & 97 & 99 \\
     & keep\,$0.8$ & 0.810 & 0.985 & 0.995 & 0.891 & +0.792 & +0.432 & 94 & 99 \\
     & keep\,$0.7$ & 0.745 & 0.975 & 0.995 & 0.849 & +0.774 & +0.423 & 87 & 97 \\
    \midrule
    \multirow{4}{*}{Crop \& resize} & keep\,$0.95$ & 0.010 & 0.065 & 0.095 & 0.049 & +0.380 & +0.011 & 1 & 2 \\
     & keep\,$0.9$ & 0.010 & 0.065 & 0.095 & 0.046 & +0.375 & +0.021 & 1 & 5 \\
     & keep\,$0.8$ & 0.005 & 0.035 & 0.060 & 0.035 & +0.392 & +0.007 & 1 & 2 \\
     & keep\,$0.7$ & 0.010 & 0.035 & 0.055 & 0.039 & +0.423 & +0.009 & 1 & 2 \\
    \bottomrule
  \end{tabular}
\end{table*}

\begin{table*}[!t]
  \centering
  \caption{%
    Ablation of extractor-input filter masks. The $26/64$ mid-band mask provides the best overall trade-off between retrieval accuracy and visual quality. Figure~\ref{fig:filter_mask} illustrates the tested masks. }
  \label{tab:v13_mask_ablation}
  \begin{adjustbox}{width=0.9\textwidth}
  \begin{tabular}{l c c c c c c c c c}
    \toprule
    & &
    \multicolumn{3}{c}{\textbf{Retrieval}} &
    \multicolumn{5}{c}{\textbf{Visual quality (LoRA\_on vs.\ LoRA\_off)}} \\
    \cmidrule(lr){3-5} \cmidrule(lr){6-10}
    Mask design & Active bins &
    Top1$\uparrow$ & MRR$\uparrow$ & $\cos_p\uparrow$ &
    PSNR$\uparrow$ & SSIM$\uparrow$ & DreamSim$\downarrow$ &
    FID$\downarrow$ & KID$\downarrow$ \\
    \midrule
    Mid-band $(3,7)$  & 30/64   & 0.805 & 0.881 & 0.789 & 22.67 & 0.694 & \textbf{0.219} & 132.32 & 0.0058 \\
    Mid-band $(4,7)$  & 26/64 & \textbf{0.860} & \textbf{0.917} & \textbf{0.802}& \textbf{22.91} & \textbf{0.704} & 0.221 & 132.05 & 0.0029 \\
    Mid-band $(5,7)$ & 21/64  & 0.780 & 0.875 & 0.786 & 22.72 & 0.697 & 0.225 & 131.63 & \textbf{0.0025} \\
    Mid-band $(5,6)$   & 13/64   & 0.790 & 0.878 & 0.789 & 22.67 & 0.697 & 0.231 & \textbf{131.52} & 0.0026 \\
    \bottomrule
  \end{tabular}
  \end{adjustbox}
\end{table*}

\begin{table*}[tp]
  \centering
  \small
  \caption{%
    Ablation of training-loss configurations. We remove the InfoNCE loss, remove the velocity anchor, align only the classifier-free guidance difference $|(v_c^\mathrm{lora} - v_u^\mathrm{lora}) - (v_c^{\mathrm{van}} - v_u^{\mathrm{van}})|$, and vary the velocity-anchor weights $(\alpha,\beta)$. Both InfoNCE and the velocity anchor are important for watermark recovery and visual fidelity. }
  \label{tab:v14_ablation}
  \begin{adjustbox}{max width=\textwidth}
  \begin{tabular}{l  c c c  c c c c c}
    \toprule
    &
    \multicolumn{3}{c}{\textbf{Retrieval}} &
    \multicolumn{5}{c}{\textbf{Visual quality (LoRA\_on vs.\ LoRA\_off)}} \\
    \cmidrule(lr){2-4} \cmidrule(lr){5-9}
    Ablation &
    Top1$\uparrow$ & MRR$\uparrow$ & $\cos_p\uparrow$ &
    PSNR$\uparrow$ & SSIM$\uparrow$ & DreamSim$\downarrow$ & FID$\downarrow$ & KID$\downarrow$ \\
    \midrule
    {AngelFingerprint} & 0.860 & 0.917 & 0.802 & 22.91 & 0.704 & 0.221 & 132.05 & 0.0029 \\
    \midrule
    w/o InfoNCE loss & 0.700 & 0.805 & 0.764 & \textbf{23.93} & \textbf{0.746} & \textbf{0.173} & \textbf{117.79} & \textbf{0.0003} \\
    w/o velocity anchor & \textbf{0.895} & \textbf{0.942} & \textbf{0.815} & 19.23 & 0.593 & 0.389 & 185.20 & 0.0366 \\
    Only align $(v_c^m - v_u^m)$ & 0.835 & 0.895 & 0.787 & 19.41 & 0.603 & 0.295 & 148.90 & 0.0151 \\
    \midrule
    Anchor $(\alpha,\beta) = (1,1)$ & 0.845 & 0.912 & 0.803 & 22.61 & 0.693 & 0.240 & 139.34 & 0.0047 \\
    Anchor $(\alpha,\beta) = (1,3)$  & 0.800 & 0.878 & 0.786 & 23.06 & 0.709 & 0.205 & 128.86 & 0.0024 \\
    Anchor $(\alpha,\beta) = (1,4)$ & 0.840 & 0.899 & 0.797 & 23.19 & 0.719 & 0.206 & 127.75 & 0.0023 \\
    Anchor $(\alpha,\beta) = (2,1)$ & 0.755 & 0.854 & 0.771 & 23.43 & 0.724 & 0.193 & 121.49 & 0.0016 \\
    \bottomrule
  \end{tabular}
  \end{adjustbox}
\end{table*}

\section{Ablation Study}

\subsection{Objective Function}
The main objective contains three terms: 
\begin{equation}
  \mathcal{L}
  = \lambda_{\mathrm{fm}}\,\mathcal{L}_{\mathrm{fm}}
  + \lambda_{\mathrm{embed}}\,\mathcal{L}_{\mathrm{embed}}
  + \lambda_{\mathrm{ref}}\,\mathcal{L}_{\mathrm{ref}} .
  \label{eq:total_}
\end{equation}
We conduct an ablation study to evaluate the effectiveness of these terms. Table~\ref{tab:v14_ablation} reports the results. First, removing the InfoNCE loss drops Top-1 from $0.860$ to $0.700$. The cosine term alone pulls each embedding toward its target but does not separate it from the others, so retrieval degrades even though visual quality slightly improves. Second, removing the velocity anchor lets the LoRA drift freely to write a visible watermark. Top-1 even rises to $0.895$, but visual quality collapses (PSNR $22.9{\to}19.2$, KID $0.0029{\to}0.0366$): the watermark becomes easy to read and easy to see. Third, aligning only the guidance difference $(v_c^m - v_u^m)$ still leaves fidelity poor (DreamSim $0.295$), which shows that both branches must be anchored separately. Next, we test different $(\alpha,\beta)$ pairs in $\mathcal{L}_{\mathrm{ref}}$. The results show that $(\alpha,\beta){=}(1,2)$ gives the best retrieval (Top-1 $0.860$), and that weighting the unconditional branch more heavily ($\beta > \alpha$) protects fidelity while keeping the mark readable.

\begin{figure}[t]
  \centering
  \includegraphics[width=0.8\columnwidth]{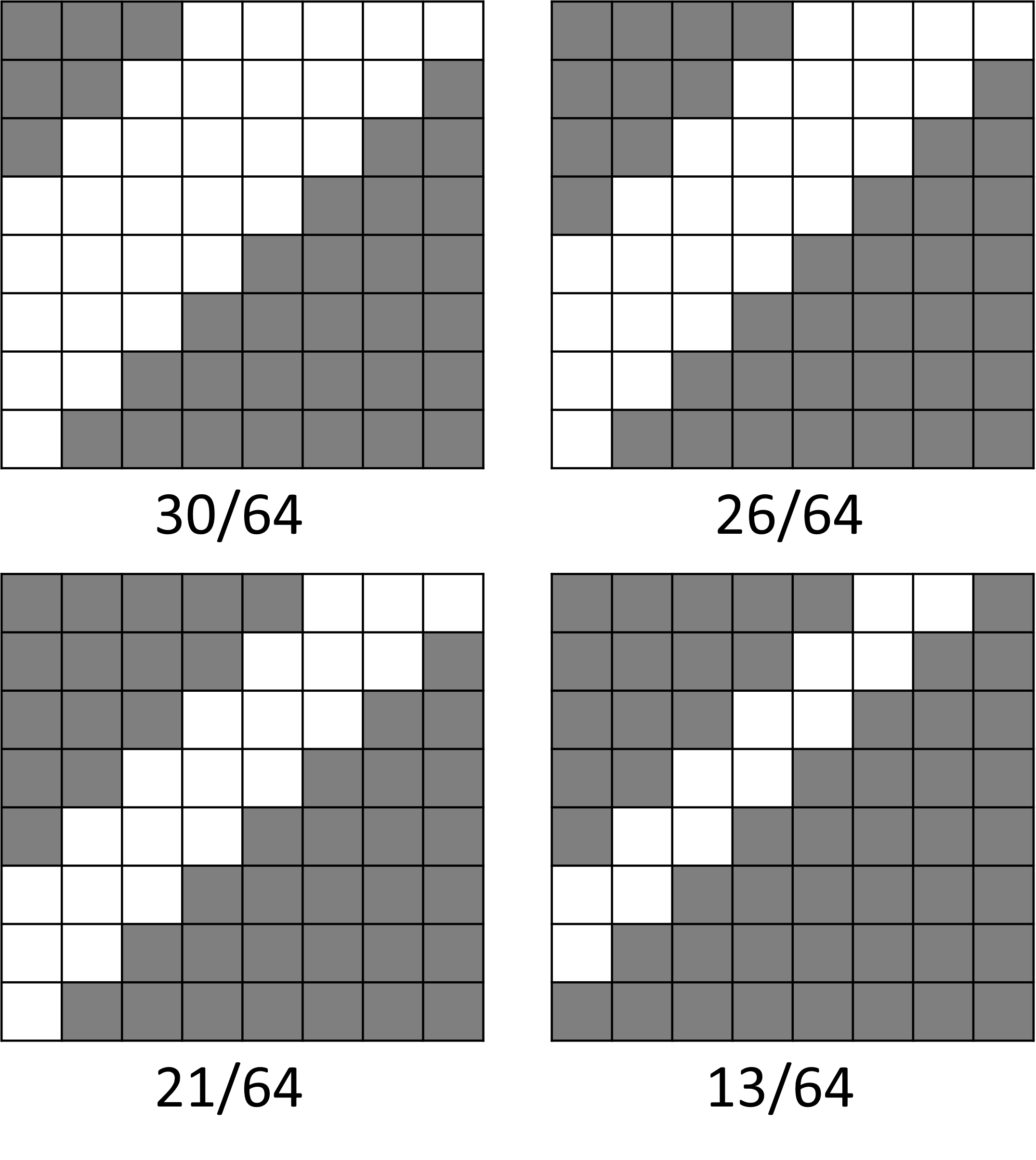}
  \caption{%
    The four JPEG band-pass filter masks tested in our ablation. A wider passband is not always better; the $26/64$ mask gives the best overall performance.}
  \label{fig:filter_mask}
\end{figure}

\subsection{Different Filter Configurations}
We test several filter configurations. Figure~\ref{fig:filter_mask} shows the four masks, and Table~\ref{tab:v13_mask_ablation} reports retrieval accuracy and visual fidelity. Interestingly, a wider passband does not necessarily improve performance; instead, the best trade-off occurs at $26/64$ active coefficients. If the passband is too wide, the extractor receives too much image content, and the gradients sent back to the LoRA become more conflicted, destabilizing joint training. If the passband is too narrow, the extractor receives too little signal and the watermark has insufficient bandwidth. The $26/64$ mask therefore provides a good balance: it passes enough mid-frequency signal for reliable extraction while regularizing the joint optimization of the LoRA and extractor.

\section{Deployment}
\label{sec:deployment}
We describe how AngelFingerprint is used end to end. A provider releases an open editing model with our watermarking LoRA merged in. Because the LoRA changes only weights, the released model is indistinguishable from an ordinary fine-tune, so a user cannot find or remove the watermark. Every edit the model makes carries an invisible record of the editing instruction. A verifier who later meets a suspect image can both detect that it came from the model and recover the instruction that produced it.

\paragraph{Release.} The watermarking LoRA is merged into a public checkpoint of the base editing model (SD3-medium or UltraEdit-SD3) and shipped through the same channels as any open-weights release. A user who inspects the model sees no change in architecture, tensor shapes, forward computation, or inference code. The LoRA is a residual update of the existing attention projections, folded into the released weights. There is no extra branch, no extra network, and no change to the sampler. The visual cost is small: the watermark shows only as a faint texture that is hard to see without close inspection, and the edit stays close to the vanilla output. The model is therefore used as an ordinary editor, and every edit it produces carries the watermark.

\paragraph{Verification.} A verifier who encounters a suspect image runs it through the extractor once and reads two signals. The first is presence: the magnitude $\lVert \hat e\rVert_2$ of the raw extractor output decides whether the image came from the watermarked model. This signal alone reaches AUC near $1$, with $100\%$ true positive rate at $1\%$ false positive rate (Section~\ref{sec:presence}).
The second is the recovered embedding: the output is a $768$-dimensional vector in CLIP-L text space, trained to lie close to the CLIP-L embedding of the edit instruction.

\paragraph{Recovering the instruction.} The embedding is a vector, not a sentence, so the verifier matches it against a pool of candidate instructions. Our design puts no constraint on how the pool is built, and the verifier can combine three sources. The first is a few hand-written hypotheses, based on which region looks manipulated, for example ``put a pond next to the elephant'' or ``add a red flag above the sign''. This is fast and often enough when the edited region is obvious. The second is a vision-language model (such as GPT-4o, Gemini, or Qwen-VL), prompted with the suspect image to enumerate many plausible instructions, so the verifier does not need to guess the edit in advance. The third is an off-the-shelf instruction corpus (MagicBrush, InstructPix2Pix, or MS-COCO captions) added as background distractors, which keeps retrieval calibrated against a realistic null pool. Each candidate is CLIP-L encoded, and the pool is ranked by cosine similarity to the recovered embedding. The top-$k$ candidates are the final output.

In addition, prior work shows that a CLIP text embedding can be inverted back to text \cite{morris2023text, zhang2025universal}. The verifier can use these methods to reconstruct the instruction directly when needed.

\paragraph{What the numbers show.} On a $200$-sentence pool, the ground-truth instruction is recovered at rank $1$ with probability $0.860$ on SD3-medium and $0.655$ on UltraEdit-SD3. These are conservative for three reasons. First, an exact phrasing is not required: a paraphrase, or any candidate that shares the CLIP-L direction of the true instruction, is usually enough for attribution. Second, the pool size is not fixed, so the verifier can expand it. Third, the content baselines (BLIP-2, PromptStealer, Florence-2, VGD) recover nothing under the same test (Top-1 $\le 0.225$ with $\Delta_{\mathrm{T1}}<0$ against the LoRA-off image), so the signal comes from the watermark, not from natural image-caption correlation.


\section{Model, Dataset, and Implementation Details}
\label{sec:setup}

\subsection{Backbone Models}

Our watermarking pipeline is designed for any diffusion model compatible with LoRA.\footnote{PEFT LoRA: \url{https://huggingface.co/docs/peft/v0.20.0/en/package_reference/lora}} We report results on two diffusion backbones: the base Stable Diffusion 3 (SD3)~\cite{esser2024scaling} and UltraEdit~\cite{zhao2024ultraedit}.

\paragraph{SD3-medium.} Our primary experiments use Stable Diffusion~3~medium~\cite{esser2024scaling}, the 2B-parameter open-weights release of Stability~AI's rectified-flow text-to-image model.\footnote{Model: \url{https://huggingface.co/stabilityai/stable-diffusion-3-medium-diffusers}} SD3 uses an MMDiT joint-attention backbone with three separate text encoders (CLIP-L, CLIP-G, T5-XXL).
In the watermarking configuration we condition on the edit instruction and use the source image as the SDEdit~\cite{meng2021sdedit} init at $\sigma{=}0.7$, i.e.\ we noise the source latent and denoise it back to a watermarked edit.

\paragraph{UltraEdit-SD3.} To evaluate whether the pipeline transfers to a purpose-built editing backbone, we swap in UltraEdit-SD3\footnote{\url{https://huggingface.co/papers/2407.05282}}~\cite{zhao2024ultraedit}, a fine-tune of SD3-medium on the UltraEdit dataset that follows the InstructPix2Pix protocol~\cite{brooks2023instructpix2pix}. Its DiT takes a $48$-channel input: three $16$-channel latents concatenated, namely the noisy latent, the VAE-encoded source image, and the VAE-encoded mask. Because we edit by instruction without a region, the mask is a blank all-white image. Generation starts from pure Gaussian noise rather than a noised source. This exercises our design under (i)~a different training distribution and (ii)~a fundamentally different sampling regime.

\subsection{Dataset: MagicBrush}

\begin{figure}[t]
  \includegraphics[width=\columnwidth]{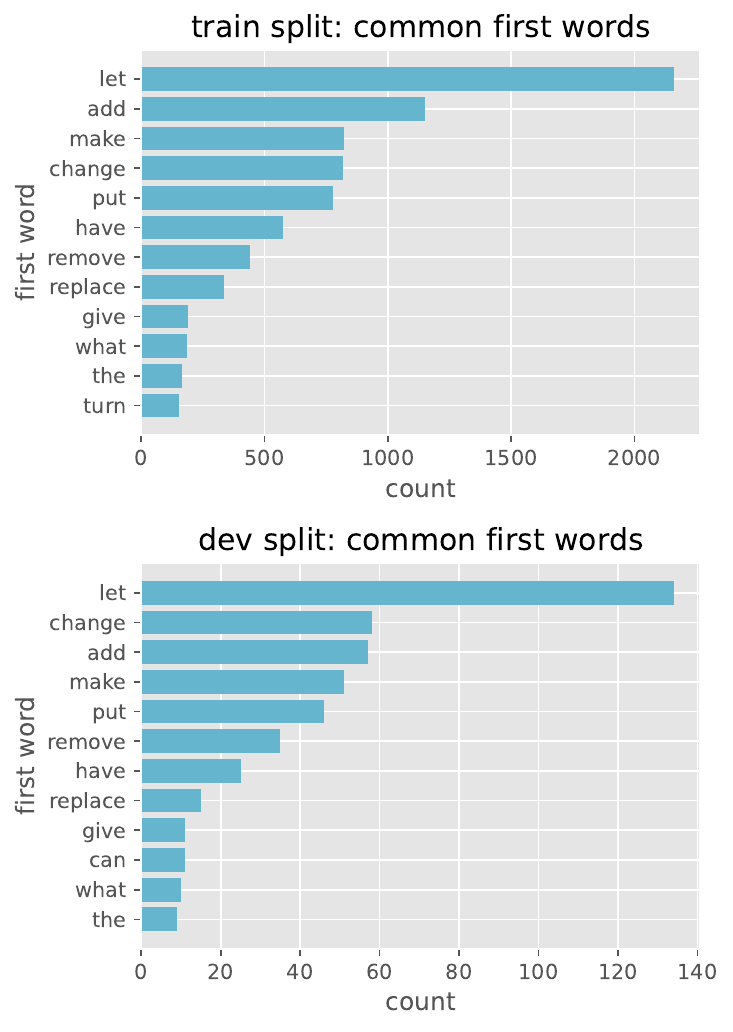}
  \caption{%
  Distribution of the first word of each editing instruction in MagicBrush. Most instructions begin with an imperative edit verb. }
  \label{fig:magicbrush_stat}
\end{figure}

We train and evaluate on the MagicBrush multi-turn image-editing dataset~\citep{zhang2023magicbrush}, an OSU-NLP corpus of human-annotated real-image edits made through the DALL-E~2 web interface. Each row is one edit turn: a source image, an instruction, a mask, and a target image. Turns are grouped into sessions (same source image, in ascending turn order); we treat each turn independently. Concrete statistics for the version we use are given in Table~\ref{tab:magicbrush}. Instructions are short (mean $6.5$ words) and are imperative edit commands; Figure~\ref{fig:magicbrush_stat} shows the distribution of their first word, which is usually an edit verb. Source images come from MS-COCO; the target image is provided but unused by our loss, since our extractor supervision comes from the CLIP-L embedding of the instruction only. For evaluation we use a fixed $N{=}200$ subset of the dev split (the first $200$ rows).

\renewcommand{\topfraction}{0.95} 

\begin{table}[t]
  \centering
  \begin{tabular}{lrr}
    \toprule
    & train & dev \\
    \midrule
    edit turns              & 8{,}807 & 528   \\
    unique sessions         & 4{,}512 & 266   \\
    mean turns / session    & 1.95    & 1.98  \\
    max turns / session     & 3       & 3     \\
    mean instruction words  & 6.5     & 6.5   \\
    median instruction words & 6      & 6     \\
    \bottomrule
  \end{tabular}
  \caption{MagicBrush statistics used in this work. We use each edit turn independently and ignore the target image; the instruction is the sole extraction target.}
  \label{tab:magicbrush}
\end{table}

\subsection{Implementation Details}

\paragraph{Extractor.} The extractor is a lightweight multi-scale CNN with five stride-$2$ convolutional blocks ($64{-}1024$ channels), global average pooling at every scale, and a $2$-layer MLP head that projects to a $768$-dimensional CLIP-L target.
Before the CNN we apply a fixed block-DCT frequency mask that keeps only the mid-band diamond $4{\le}i{+}j{\le}7$ ($26$ of $64$ coefficients per $8{\times}8$ block) and zeros the rest. The mask is a fixed binary band with no learnable parameters, which prevents the extractor from reading low-frequency content directly.

\paragraph{Watermarking loss.} The LoRA adapter is jointly trained with the extractor to minimize the weighted sum of loss terms defined in the main paper: (1) the flow-matching loss $\mathcal{L}_{\mathrm{fm}}$ with weight $\lambda_{\mathrm{fm}}{=}0.5$; (2) the embedding-recovery loss $\mathcal{L}_{\mathrm{embed}}$ with weight $\lambda_{\mathrm{embed}}{=}2.0$, combining a cosine alignment term and an InfoNCE term (temperature $\kappa{=}0.07$); and (3) the $4$-point velocity-anchor loss $\mathcal{L}_{\mathrm{ref}}$, which regresses the LoRA-modified conditional and unconditional velocities to their vanilla-DiT counterparts with branch weights $\alpha{=}1.0$ and $\beta{=}2.0$. Both the DiT LoRA and the extractor are updated end-to-end.

\paragraph{LoRA configuration.} We use LoRA rank $r{=}96$ and \texttt{lora\_alpha}${=}48$. For SD3-medium we adapt the attention query, key, value, and output projections (\texttt{to\_q}, \texttt{to\_k}, \texttt{to\_v}, \texttt{to\_out}) in every DiT block. For UltraEdit-SD3 (48-channel DiT) we additionally adapt the MMDiT text-stream projections (\texttt{add\_q\_proj}, \texttt{add\_k\_proj}, \texttt{add\_v\_proj}, \texttt{to\_add\_out}). Each matching linear layer receives its own LoRA adapter.

\paragraph{Training.} AdamW with base learning rate $1{\times}10^{-4}$, $\text{bf16}$-mixed precision, batch size $4$ (SD3-medium). Training uses single-step flow matching with logit-normal timestep sampling clamped to $[0.05,\,0.95]$. We train for $30$ epochs.

\paragraph{Deployment protocols.} We evaluate under two deployment regimes, one per backbone. The SD3-medium regime runs SDEdit-style img2img at $640{\times}640$ with $14$ Euler steps, CFG scale $3.5$, strength $0.7$, and no negative prompt; this is our default for all SD3-medium results. The UltraEdit regime runs UltraEdit-SD3 through its native InstructPix2Pix pipeline at $512{\times}512$ with $14$ steps, text CFG $3.5$, image CFG $1.0$, and a pure-noise start. In both cases we generate matched pairs at LoRA scales $\{0, 1\}$ using identical seeds so that per-pair perceptual metrics (LPIPS, PSNR, SSIM, DreamSim) isolate the watermark's marginal cost.

\paragraph{Hardware and wall-clock.} All training runs use a single NVIDIA RTX~PRO~5000 GPU (48\,GB) and take approximately $40$ hours per configuration for the full $30$ epochs. No distributed or multi-GPU training was used; all reported ablations were run sequentially on the same GPU.

\section{Traceability: Confirming Origins}
\label{sec:presence}
AngelFingerprint recovers the CLIP embedding of the editing instruction. A verifier may first ask a simpler question: given an image, can we confirm that it was produced by our model, before recovering the embedding? The answer is yes, and the check needs no extra model. This is a provenance check, the same goal that prior watermarks pursue with a fixed ID. In our setting the extractor is private and available only to the verifier. We show that the same extractor is not only a prompt recoverer but also a watermark-presence verifier.

The signal is the output magnitude. Let $\hat e = E_\phi(H(x))$ be the raw extractor output on image $x$, before the $\ell_2$ normalization we use for retrieval. We simply measure $\lVert \hat e\rVert_2$.

\subsection{Setup}
We build three groups of $200$ images and pass all of them through the same extractor. Group~A is watermarked: our model with the LoRA on. Group~B is the same pipeline with the LoRA disabled, so it is the same generator running the same edit without our watermark. This is the closest realistic negative an adversary could produce. Group~C is raw natural photographs, the MagicBrush source images, never produced by any generator. For each image we compute the output norm $\lVert \hat e\rVert_2$. We score how well each separates group~A from the negatives B+C by ROC-AUC.

\subsection{The Output Norm Is a Perfect Detector}
\begin{figure}[t]
  \includegraphics[width=\columnwidth]{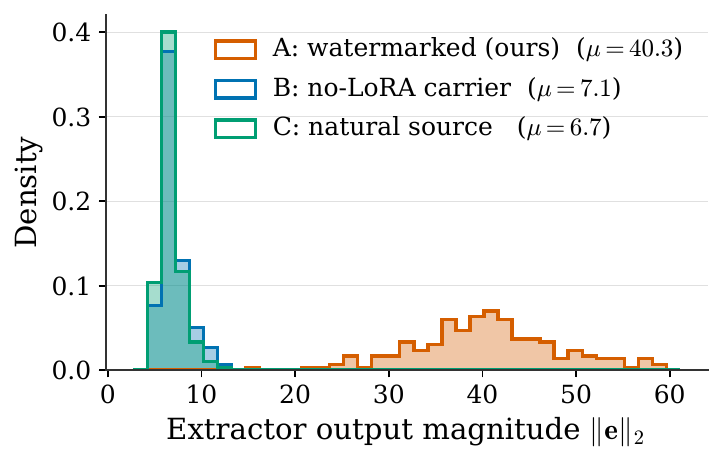}
  \caption{%
  Distribution of extractor-output norms for watermarked images, LoRA-off images, and natural photographs. }
  \label{fig:confirming}
\end{figure}

The output norm separates watermarked from unwatermarked images almost perfectly (Table~\ref{tab:presence_auc}), achieving AUC $1.000$ in every comparison. Figure~\ref{fig:confirming} illustrates the extractor-output norm distribution; the watermarked group A is clearly separated from both group B and group C. Table~\ref{tab:presence_l2} reports the exact statistics.

\begin{table}[t]
  \centering 
  \caption{Watermark presence detection. ROC-AUC of the extractor-output norm when separating watermarked images (A) from unwatermarked negatives: the same pipeline with the LoRA off (B), natural photographs (C), and their union (B+C).}
  \label{tab:presence_auc}
  \begin{tabular}{lccc}
    \toprule
    Signal & A vs B & A vs C & A vs B+C \\
    \midrule
    $\lVert \hat e\rVert_2$       & {1.0000} & {1.0000} & {1.0000} \\
    \bottomrule
  \end{tabular}
\end{table}

\begin{table}[t]
  \centering 
  \caption{Distribution of the extractor-output norm $\lVert \hat e\rVert_2$ across the three groups. The watermarked and unwatermarked ranges do not overlap: the $5$th percentile of A is about three times the $95$th percentile of B.}
  \label{tab:presence_l2}
  \begin{tabular}{lrrrr}
    \toprule
    Group & mean & std & 5th \% & 95th \% \\
    \midrule
    A (watermarked) & \textbf{40.34} & 7.69 & 27.82 & 53.79 \\
    B (LoRA off)    & 7.06 & 1.43 & 5.41 & 10.08 \\
    C (natural)     & 6.73 & 1.19 & 5.31 & 8.92 \\
    \bottomrule
  \end{tabular}
\end{table}

Watermarked images have a mean norm of $40.3$, while both negatives sit near $7$. The gap is complete: the $5$th percentile of the watermarked group ($27.8$) is still about three times the $95$th percentile of the LoRA-off group ($10.1$). The two ranges do not overlap, so the AUC reaches the ceiling.

This phenomenon is a natural byproduct of the cosine embedding loss. The loss optimizes the extractor to amplify the specific feature direction encoding the watermark. Consequently, the extractor exhibits strong activations when the LoRA signal is present.

\subsection{A One-Line Detector}
Because the norm alone separates the groups, a single threshold detects the watermark. At $\lVert \hat e\rVert_2 > 12$ we reach a $100\%$ true positive rate at a $1\%$ false positive rate on this test. The norm needs no candidate prompt, so it works even when the verifier has nothing to test against. Since only the verifier holds the extractor, this check is available to the verifier but not to an adversary without it.

\subsection{Limitations of This Confirmation}
We report this as an empirical property, with three limits. First, the norm scale is measured on our SD3-medium model, so a different backbone may have a different baseline and the threshold should be re-calibrated per model. Second, our negatives are same-generator LoRA-off images and natural photographs; stronger adversarial negatives, such as other diffusion outputs or images perturbed to spoof the norm, are not covered. Third, $200$ images per group are enough to show AUC near $1$ but not to bound the extreme tail, so a target false positive rate below $10^{-4}$ would need a larger negative set. A trained presence head with hard negatives could harden the detector against spoofing, which we leave for future work.



\section{Generative AI Usage Statement}
During the preparation of this work, the authors utilized Large Language Models, such as ChatGPT and Gemini, to improve the readability, grammar, and fluency of the writing. This editorial assistance was applied throughout the manuscript, including the main text, figure captions, and table descriptions. The authors reviewed all AI-suggested edits and take full responsibility for the content of the paper.



\end{document}